\documentclass[10pt,twocolumn,letterpaper]{article}

\usepackage{cvpr}              

\usepackage{graphicx}
\usepackage{amsmath}
\usepackage{amssymb}
\usepackage{booktabs}
\usepackage{multirow}

\usepackage[pagebackref,breaklinks,colorlinks]{hyperref}
\usepackage{lipsum}
\newcommand\blfootnote[1]{%
\begingroup
\renewcommand\thefootnote{}\footnote{#1}%
\addtocounter{footnote}{-1}%
\endgroup
}

\usepackage[capitalize]{cleveref}
\crefname{section}{Sec.}{Secs.}
\Crefname{section}{Section}{Sections}
\Crefname{table}{Table}{Tables}
\crefname{table}{Tab.}{Tabs.}

\def\cvprPaperID{6385} 
\def\confName{CVPR}
\def\confYear{2022}

\begin{document}

\title{Frame-wise Action Representations for Long Videos via\\ Sequence Contrastive Learning}

\author{Minghao Chen$^{1*}$  \quad Fangyun Wei$^{2\dagger}$\quad Chong Li$^{2}$ \quad Deng Cai$^{1}$  \vspace{4pt}\\
	$^1$State Key Lab of CAD\&CG, College of Computer Science, Zhejiang University \\
    $^2$Microsoft Research Asia \\
    {\tt\small minghaochen01@gmail.com} \quad
	{\tt\small \{fawe, chol\}@microsoft.com} \quad {\tt\small dengcai@cad.zju.edu.cn} \\
}

\maketitle
\begin{abstract}
    Prior works on action representation learning mainly focus on designing various architectures to extract the global representations for short video clips. In contrast, many practical applications such as video alignment have strong demand for learning dense representations for long videos. In this paper, we introduce a novel contrastive action representation learning (CARL) framework to learn frame-wise action representations, especially for long videos, in a self-supervised manner. Concretely, we introduce a simple yet efficient video encoder that considers spatio-temporal context to extract frame-wise representations. Inspired by the recent progress of self-supervised learning, we present a novel sequence contrastive loss (SCL) applied on two correlated views obtained through a series of spatio-temporal data augmentations. SCL optimizes the embedding space by minimizing the KL-divergence between the sequence similarity of two augmented views and a prior Gaussian distribution of timestamp distance. Experiments on FineGym, PennAction and Pouring datasets show that our method outperforms previous state-of-the-art by a large margin for downstream fine-grained action classification. Surprisingly, although without training on paired videos, our approach also shows outstanding performance on video alignment and fine-grained frame retrieval tasks. Code and models are available at \url{https://github.com/minghchen/CARL_code}.
    
\end{abstract}
\blfootnote{*Accomplished during Minghao Chen’s internship at MSRA.}
\blfootnote{ $^{\dagger}$Corresponding author.}
\vspace{-4mm}
\section{Introduction}
\label{intro}
\begin{figure}[t]
  \centering
  \begin{subfigure}{\linewidth}
    \centering
    \includegraphics[width=0.87\linewidth]{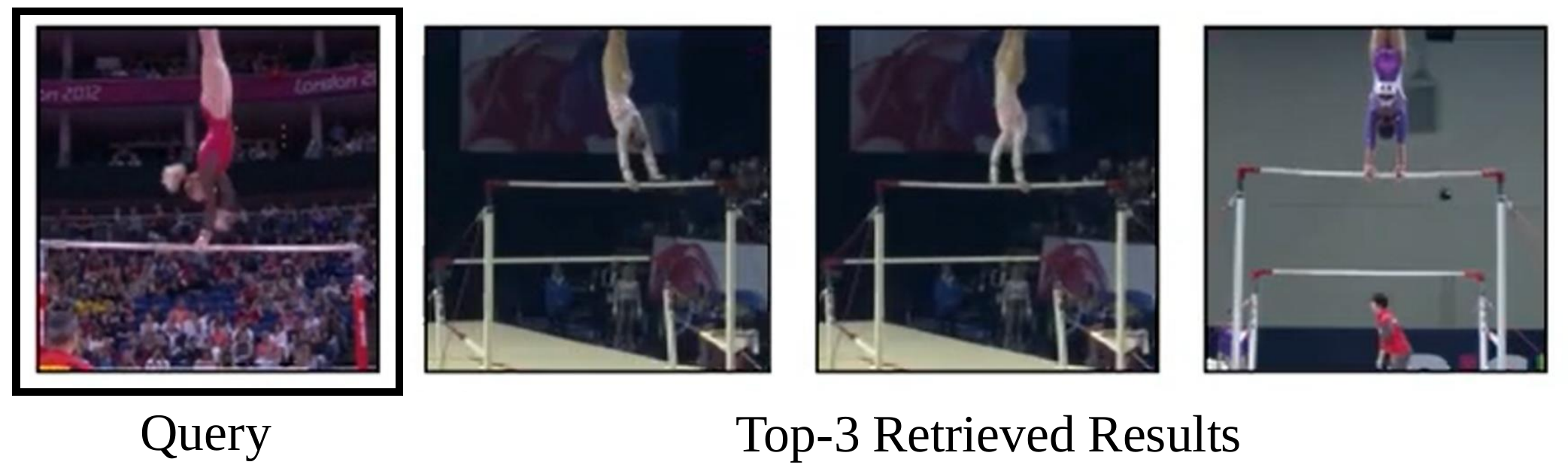}
    \caption{Fine-grained frame retrieval on FineGym dataset.}
  \end{subfigure}
  \begin{subfigure}{\linewidth}
    \centering
    \includegraphics[width=0.86\linewidth]{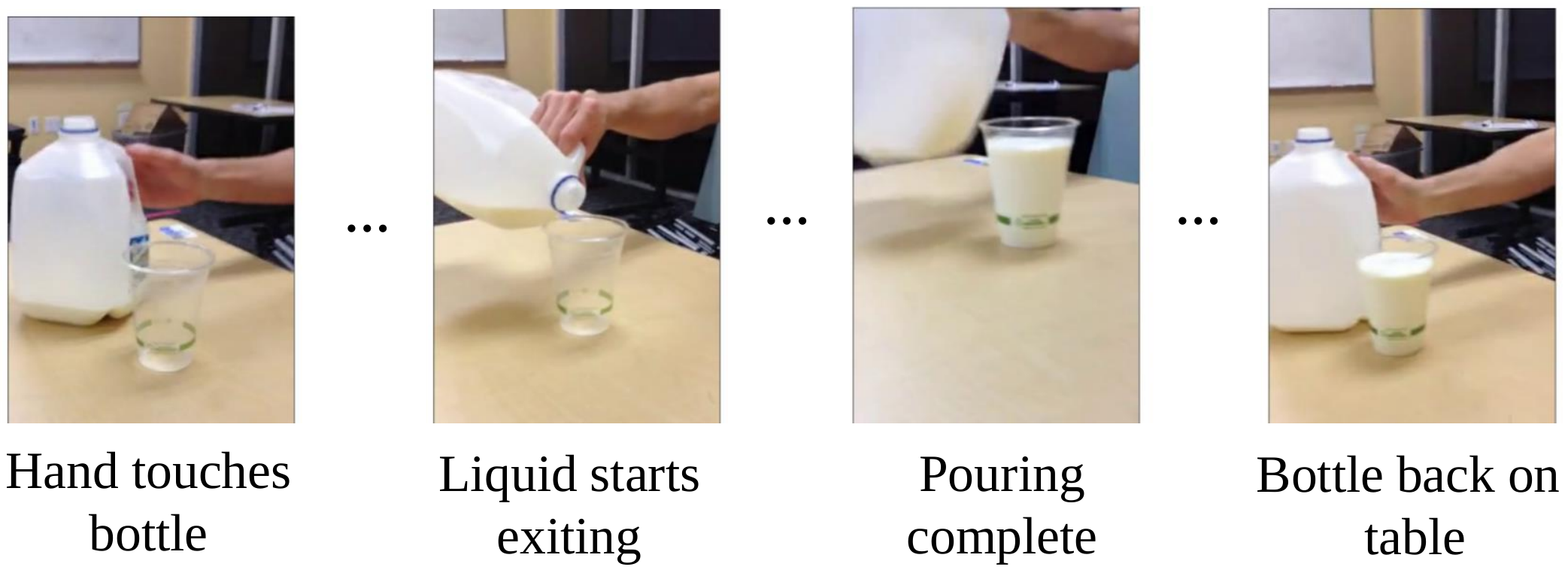}
    \caption{Phase boundary detection on Pouring dataset.}
  \end{subfigure}
  \begin{subfigure}{\linewidth}
    \centering
    \includegraphics[width=0.85\linewidth]{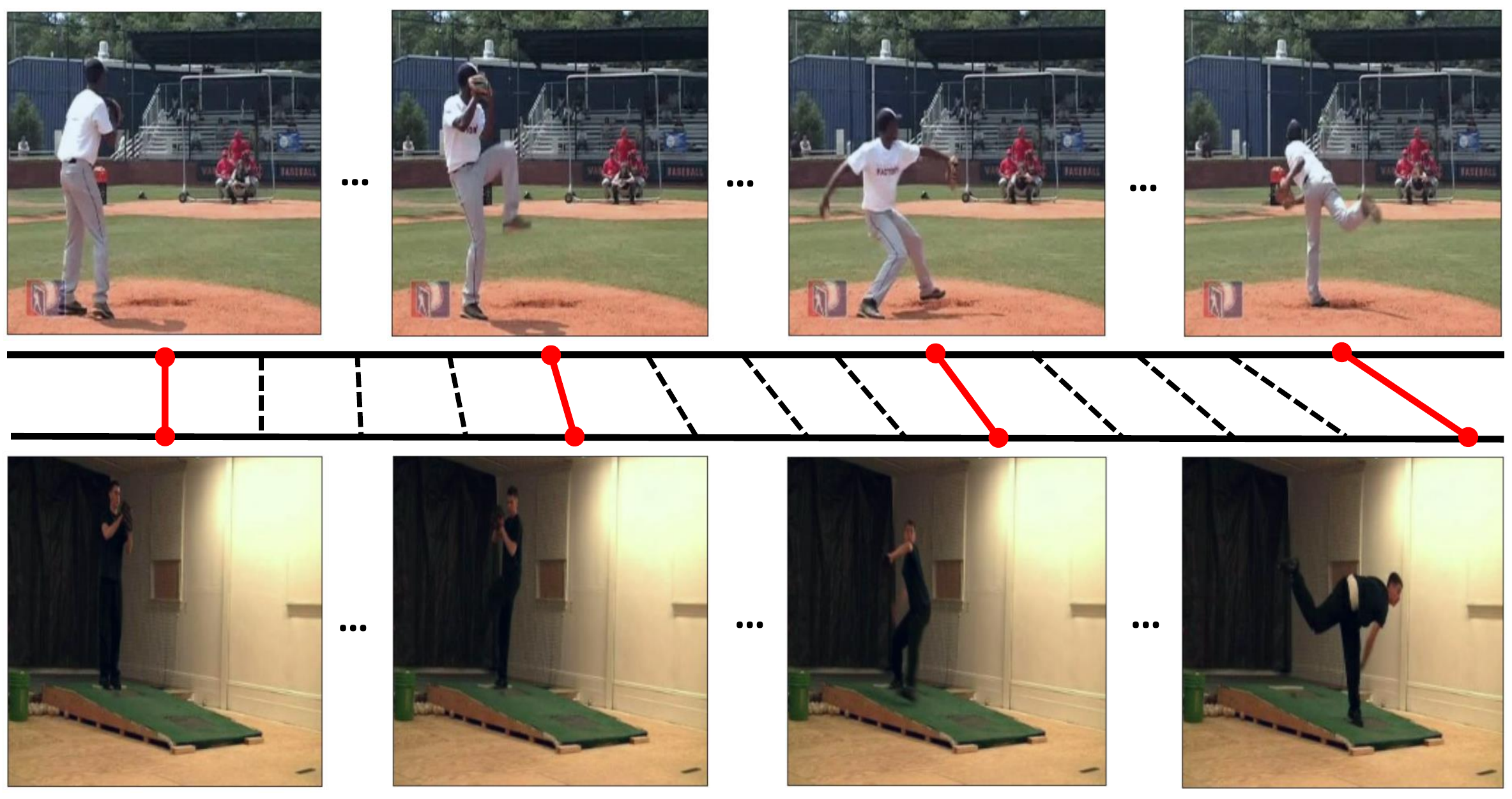}
    \caption{Temporal video alignment on PennAction dataset.}
  \end{subfigure}
  \vspace{-0.2cm}
  \caption{Multiple applications of our frame-wise representation learning on various datasets:
  (a) Fine-grained frame retrieval on FineGym~\cite{FineGym}. (b) Phase boundary detection on Pouring~\cite{TCN}. (c) Temporal video alignment on PennAction~\cite{PennAction}. As shown in the figures, the representations obtained through our method (CARL) are invariant to the appearance, viewpoint and background.
  }
   \vspace{-0.2cm}
   \label{fig1}
\end{figure}

In the last few years, deep learning for video understanding~\cite{TwoStream, C3D, Kinetics, Nonlocal, SlowFast, VTN, ViViT,xu2021cross} has achieved great success on video classification task~\cite{UCF101,Kinetics,Something-Something}. Networks such as I3D~\cite{Kinetics} and SlowFast~\cite{SlowFast} always take short video clips (e.g., 32 frames or 64 frames) as input and extract global representations to predict the action category. In contrast, many practical applications, e.g., sign language translation~\cite{camgoz2018neural,camgoz2020sign,chen2022simple}, robotic imitation learning~\cite{Liu2018ImitationFO, TCN}, action alignment~\cite{LAV,GTA,TemporalAlignment} and phase classification~\cite{PennAction, TCC, FineGym, Breakfast} require algorithms having ability to model long videos with hundreds of frames and extract frame-wise representations rather than the global features (Fig.~\ref{fig1}).

Previous methods~\cite{Breakfast, Cooking, FineGym} have made an effort to learn frame-wise representations via supervised learning, where sub-actions or phase boundaries are annotated. However, it is time-consuming and even impractical to manually label each frame and exact action boundaries~\cite{GTA} on large-scale datasets, which hinders the generalization of models trained with fully supervised learning in realistic scenarios. To reduce the dependency of labeled data, some methods such as TCC~\cite{TCC}, LAV~\cite{LAV} and GTA~\cite{GTA} explored weakly-supervised learning by using either cycle-consistency loss~\cite{TCC} or soft dynamic time warping~\cite{LAV, GTA}. All these methods rely on video-level annotations and the training is conducted on the paired videos describing the same action. This setting obstructs them from applying on more generic video datasets where no labels are available.

The goal of this work is to learn frame-wise representations with spatio-temporal context information for long videos in a self-supervised manner. Inspired by the recent progress of contrastive representation learning~\cite{SimCLR, MoCoV2, BYOL, SwAV}, we present a novel framework named contrastive action representation learning (CARL) to achieve our goal. We assume no labels are available during training, and videos in both training and testing sets have long durations (hundreds of frames). Moreover, we do not rely on video pairs of the same action for training. Thus it is practical to scale up our training set with less cost. 

Modeling long videos with hundreds of frames is challenging. It is non-trivial to directly use off-the-shelf backbones designed for short video clip classification, since our task is to extract frame-wise representations for long videos. In our work, we present a simple yet efficient video encoder that consists of a 2D network to encode spatial information per frame and a Transformer~\cite{Transformer} encoder to model temporal interaction. The frame-wise features are then used for representation learning.

Recently, SimCLR~\cite{SimCLR} uses instance discrimination~\cite{wu2018unsupervised} as the pretext task and introduces a contrastive loss named NT-Xent, which maximizes the agreement between two augmented views of the same data. In their implementation, all instances other than the positive reference are considered as negatives. Unlike image data, videos provide more abundant instances (each frame is regarded as an instance), and the neighboring frames have high semantic similarities. Directly regarding these frames as negatives may hurt the learning. To avoid this issue, we present a novel sequence contrastive loss (SCL), which optimizes the embedding space by minimizing the KL-divergence between the sequence similarity of two augmented video views and a prior Gaussian distribution.

The main contributions of this paper are summarized as follows:
\begin{itemize}
\item We propose a novel framework named contrastive action representation learning (CARL) to learn frame-wise action representations with spatio-temporal context information for long videos in a self-supervised manner. Our method does not rely on any data annotations and has no assumptions on datasets.

\item  We introduce a Transformer-based network to efficiently encode long videos and a novel sequence contrastive loss (SCL) for representation learning. Meanwhile, a series of spatio-temporal data augmentations are designed to increase the variety of training data.

\item Our framework outperforms the state-of-the-art methods by a large margin on multiple tasks across different datasets. For example, under the linear evaluation protocol on FineGym~\cite{FineGym} dataset, our framework achieves 41.75\% accuracy, which is +13.94\% higher than the existing best method GTA~\cite{GTA}. On PennAction~\cite{PennAction} dataset, our method achieves 91.67\% for fine-grained classification, 99.1\% for Kendall’s Tau, and 90.58\% top-5 accuracy for fine-grained frame retrieval, which all surpass the existing best methods.
\end{itemize}
\section{Related Works}
\begin{figure*}[htp!]
    \centering
    \includegraphics[width=0.99\textwidth]{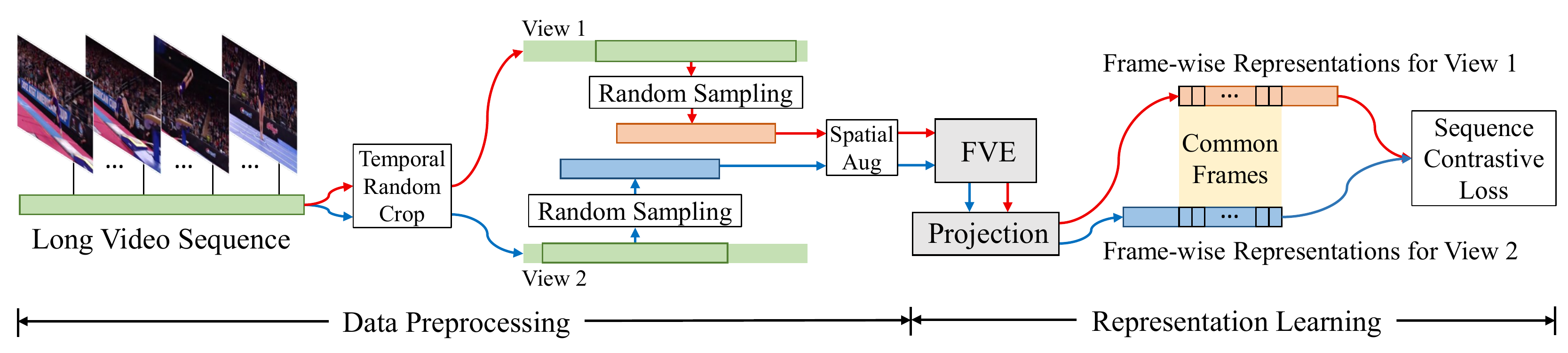}
    \caption{Overview of our framework (CARL). Two augmented views are constructed from a training video through a series of spatio-temporal data augmentations. The frame-level video encoder (FVE) and the projection head are optimized by minimizing the proposed sequence contrastive loss (SCL) between two views.}
    \vspace{-3mm}
    \label{fig2}
\end{figure*}
\noindent\textbf{Conventional Action Recognition.}
Various challenging video datasets~\cite{UCF101, ActivityNet, Charades, Kinetics, MomentsIT} have been constructed to reason deeply about diverse scenes and situations. These datasets provide labels of high-level concepts or detailed physical aspects for short videos or trimmed clips. To tackle video
recognition, large amounts of architectures have been proposed~\cite{TwoStream, TSN, C3D, Kinetics, SlowFast, Nonlocal, VTN, TimeSformer, ViViT}. Most networks are based on 3D Convolution layers and combined with the techniques in image recognition~\cite{C3D, Kinetics, SlowFast}, e.g., residual connections~\cite{ResNet} and ImageNet pre-training~\cite{ImageNet}. Some works~\cite{Nonlocal, VTN} find that 3D ConvNets have insufficient receptive fields and become the bottleneck of the computational budget. 

Recently, Transformers~\cite{Transformer} achieved great success in the field of computer vision, e.g., ViT~\cite{ViT} and DETR~\cite{DETR}. There are also several works that extend Transformers to video recognition, such as TimeSformer~\cite{TimeSformer} and ViViT~\cite{ViViT}. Due to the strong capacity of Transformers and the global receptive field, these methods have become new state-of-the-art. Combining 2D backbones and Transformers, VTN~\cite{VTN} can efficiently process long video sequences. However, these architecture are all designed for video classification and predict one global class for a video. 

\noindent\textbf{Fine-grained Action Recognition.}
There are also some datasets~\cite{Breakfast, Cooking, PennAction, FineGym} that investigate fine-grained action recognition. They decompose an action into some action units, sub-actions, or phases. As a result, each video contains multiple simple stages, e.g., wash the cucumber, peel the cucumber, place the cucumber, take a knife, and make a slice in preparing cucumber~\cite{Cooking}. However, these fine-level labels are more expensive to collect, resulting in a limited size of these datasets. GTA~\cite{GTA} argues that these boundary of manual annotations are subjective. Therefore, self-supervised learning for fine-level representations is a promising direction.

\noindent\textbf{Self-supervised Learning in Videos.}
Previous methods of self-supervised learning in videos construct pretext tasks, including inferring the future~\cite{DenseCode}, discriminating shuffled frames~\cite{SAL} and predicting speed~\cite{SpeedNet}. There are also some alignment-based methods, where a pair of videos are trained with cycle-consistent loss~\cite{TCC} or soft dynamic time warping (DTW)~\cite{LAV, D3TW, GTA}.
Recently, the contrastive learning methods~\cite{SimCLR, MoCoV2, BYOL,wei2021aligning} based on instance discrimination have shown superior performance on 2D image tasks. Some works~\cite{TCN, VideoConstrast, largescale, SeCo, VideoCL} also use this contrastive loss for video representation learning. They treat different frames in a video~\cite{TCN, SeCo, VideoCL} or different clips~\cite{VideoConstrast, largescale} in other videos as negative samples. Different from these methods, our goal is fine-grained temporal understanding of videos and we treat a long sequence of frames as input data. The most relevant work to ours is \cite{li2021towards}, which utilizes 3D human keypoints for self-supervised acton discovery in long kinematic videos.
\vspace{-1mm}
\section{Method}
\vspace{-1mm}
In this section, we introduce a novel framework named contrastive action representation learning (CARL) to learn frame-wise action representations in a self-supervised manner. In particular, our framework is designed to model long video sequences by considering spatio-temporal context. We first present an overview of the proposed framework in Section~\ref{sec:overview}. Then we introduce the details of view construction and data augmentation in Section~\ref{sec:view}. Next, we describe our frame-level video encoder in Section~\ref{sec:encoder}. Finally, the proposed sequence contrastive loss (SCL) and its design principles are introduced in Section~\ref{sec:loss}.

\subsection{Overview}
\label{sec:overview}
Figure~\ref{fig2} displays an overview of our framework. We first construct two augmented views for an input video through a series of spatio-temporal data augmentations. This step is named data preprocessing. Then we feed two augmented views into our frame-level video encoder (FVE) to extract dense representations. Following SimCLR~\cite{SimCLR}, FVE is appended with a small projection network which is a two-layer MLP for obtaining latent embeddings. Due to the fact that temporally adjacent frames are highly correlated, we assume that the similarity distribution between two augmented views should follow a prior Gaussian distribution. Based on the assumption, we propose a novel sequence contrastive loss (SCL) to optimize frame-wise representations in the embedding space.

\subsection{View Construction}
\label{sec:view}
We first introduce the view construction step of our method, as shown in the `data preprocessing' part in Figure~\ref{fig2}. Data augmentation is crucial to avoid trivial solutions in self-supervised learning~\cite{SimCLR, MoCoV2}. Different from prior methods designed for image data which only require spatial augmentations, we introduce a series of spatio-temporal data augmentations to further increase the variety of videos.

Concretely, for a training video $\boldsymbol{V}$ with $S$ frames, we aim to construct two augmented videos with $T$ frames independently through a series spatio-temporal data augmentations. For temporal data augmentation, we first perform temporal random crop on $\boldsymbol{V}$ to generate two randomly cropped clips with the length of $[T,\alpha T]$ frames, where $\alpha$ is a hyper-parameter controlling maximum crop size. During this process, we guarantee at least $\beta$ percent of overlapped frames existing between two clips. Then we randomly sample $T$ frames for each video sequence, and obtain $\boldsymbol{V}^1=\{\boldsymbol{v}_i^1~|~1\leq i \leq T\}$, and $\boldsymbol{V}^2=\{\boldsymbol{v}_i^2~|~1\leq i \leq T\}$, where ${\boldsymbol{
v}}_i^1$ and ${\boldsymbol{v}}_i^2$ represent $i$-th frame from $\boldsymbol{V}^1$ and $\boldsymbol{V}^2$, respectively. We set $T=240$ by default. For the videos with less than $T$ frames, empty frames are padded before cropping. Finally, we apply several temporal-consistent spatial data augmentations, including random resize and crop, horizontal flip, random color distortions, and random Gaussian blur, on $\boldsymbol{V}^1$ and $\boldsymbol{V}^2$ independently.

\begin{figure}[t]
    \centering
    \includegraphics[width=0.9\linewidth]{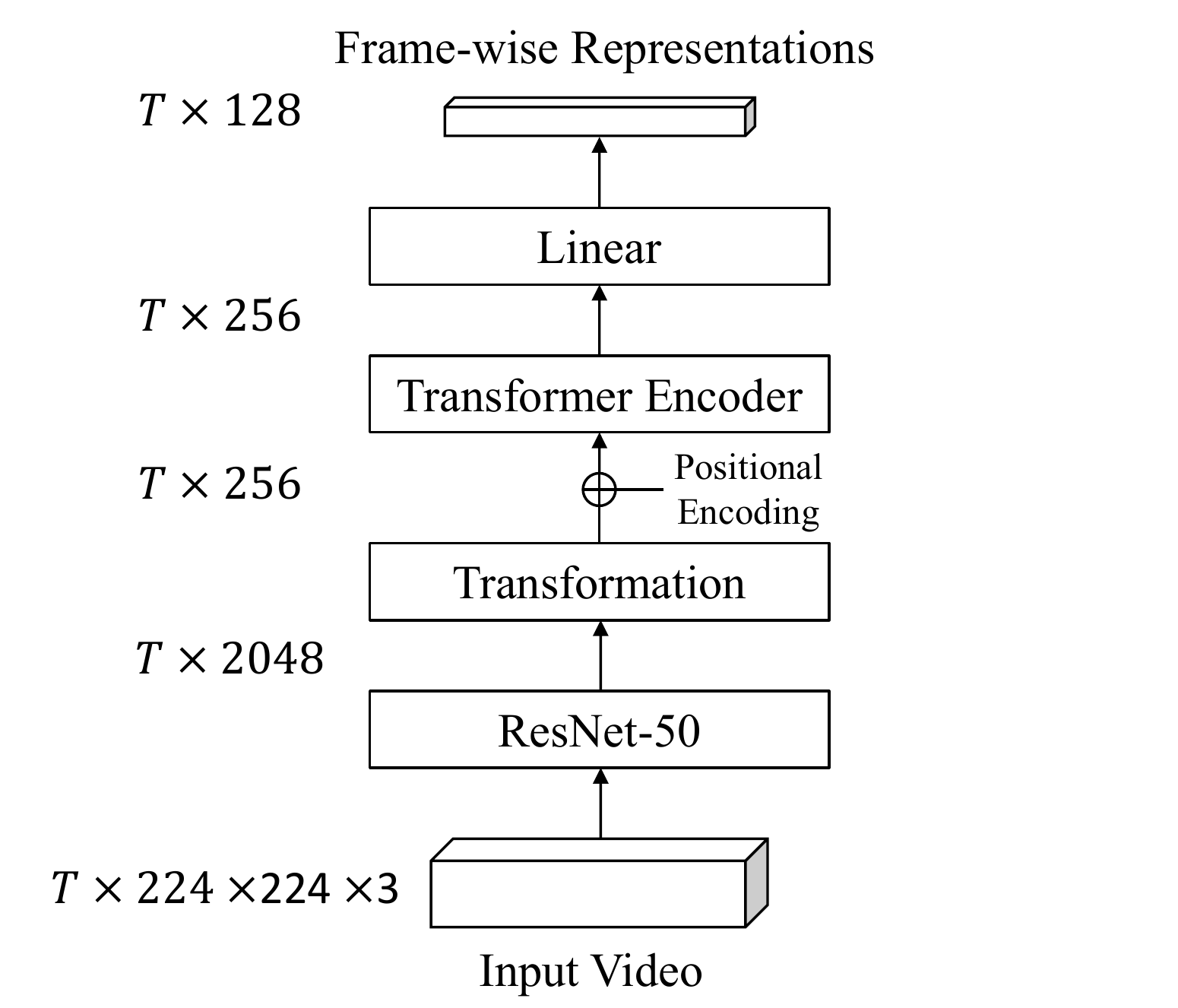}
    \caption{Architecture of the proposed  frame-level video encoder (FVE). The input is a long video with $T$ frames and the outputs are frame-wise representations. ResNet-50 is pre-trained on ImageNet. We freeze the first four residual blocks of ResNet-50 and only finetune the last block.}
    \label{fig3}
    \vspace{-4mm}
\end{figure}

\subsection{Frame-level Video Encoder}
\label{sec:encoder}
It is non-trivial to directly apply video classification backbones~\cite{C3D, Kinetics, SlowFast} to model long video sequences with hundreds of frames due to the huge computational cost. TCC~\cite{TCC} presents a video encoder that combines 2D ResNet and 3D Convolution to generate frame-wise features. However, stacking too many 3D Convolutional layers leads to unaffordable computational costs. As a result, this kind of design may have limited receptive fields to capture temporal context. Recently, Transformers~\cite{Transformer} achieved great progress in computer vision~\cite{ViT, DETR}. Transformers utilize the attention mechanism to solve sequence-to-sequence tasks while handling long-range dependencies with ease. In our network implementation, we adopt the Transformer encoder as an alternative to model temporal context.

Figure~\ref{fig3} shows our frame-level video encoder (FVE). To seek the tradeoff between representation performance and inference speed, we first use a 2D network, e.g., ResNet-50~\cite{ResNet}, along temporal dimension to extract spatial features for the RGB video sequence of size ${T \times 224 \times 224 \times 3}$. Then a transformation block that consists of two fully connected layers with batch normalization and ReLU is applied to project the spatial features to the intermediate embeddings of size $T \times 256$. Following common practice, we add the sine-cosine positional encoding~\cite{Transformer} on top of the intermediate embeddings to encode the order information. Next, the encoded embeddings are fed into the 3-layer Transformer encoder to model temporal context. At last, a linear layer is adopted to obtain the final frame-wise representations $\boldsymbol{H} \in \mathbb{R}^{T\times128}$. We use $\boldsymbol{h}_i$ ($1\leq i \leq T$) to denote the representation of $i$-th frame.

The 2D ResNet-50 network is pre-trained on ImageNet~\cite{ImageNet}. Considering the limited computational budget, we freeze the first four residual blocks since they already learned favorable low-level visual representations by pretraining. This simple design ensures that our network can be trained and tested on videos with more than 500 frames. VTN~\cite{VTN} adopt a similar hybrid Transformer-based network to perform video classification. They use the $[C L S]$ token to generate a global feature, while our network is designed to extract frame-wise representations by considering the spatio-temporal context. In addition, our network explores modeling much more prolonged video sequences.

\begin{figure}
    \centering
    \includegraphics[width=0.8\linewidth]{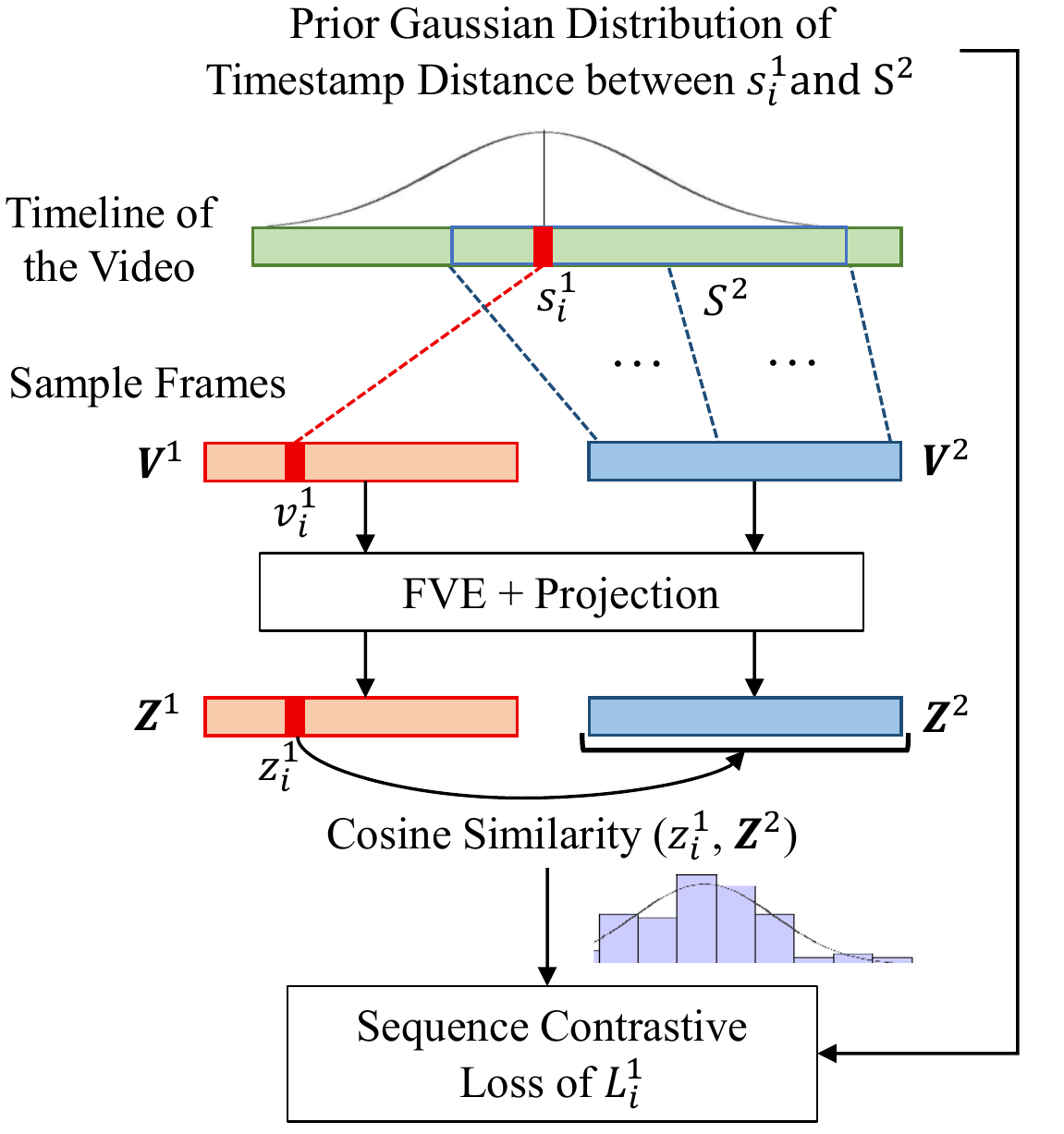}
    \caption{Illustration of the proposed sequence contrastive loss. We use the loss computation of $\boldsymbol{v}_i^1\in \boldsymbol{V}^1$ as the example. We first compute a prior Gaussian distribution of timestamp distance ($s_i^1 - s_1^2,\cdots, s_i^1 - s_T^2$). Then the embedding similarity distribution between $\boldsymbol{z}_i^1$ and $\boldsymbol{Z}^2$ is calculated. We minimize the KL-divergence of two distributions in the embedding space.}
    \label{fig4}
    \vspace{-3mm}
\end{figure}

\subsection{Sequence Contrastive Loss}
\label{sec:loss}
SimCLR~\cite{SimCLR} introduces a contrastive loss named NT-Xent by maximizing agreement between augmented views of the same instance. 

Unlike self-supervised learning for images, videos provide abundant sequential information, which is a vital supervisory signal.
For typical instance discrimination, all instances other than the positive reference are considered as negatives. However, the neighboring frames around the reference frame are highly correlated. Directly regarding these frames as negatives may hurt the learning. Learning principles should be carefully designed to avoid this issue. To optimize frame-wise representations, we propose a novel sequence contrastive loss (SCL) which minimizes the KL-divergence between the embedding similarity of two augmented views and the prior Gaussian distribution, as shown in Figure~\ref{fig4}.

Concretely, following SimCLR, we use a small projection network $g(\cdot)$ which is a two-layer MLP to project frame-wise representations $\boldsymbol{H}$ encoded by the proposed FVE to the latent embeddings $\boldsymbol{Z} = g(\boldsymbol{H})$. Let $\boldsymbol{Z}^1=\{\boldsymbol{z}_i^1~|~1\leq i \leq T\}$ and $\boldsymbol{Z}^2=\{\boldsymbol{z}_i^2~|~1\leq i \leq T\}$ denote the latent embeddings of $\boldsymbol{V}^1$ and $\boldsymbol{V}^2$, where $\boldsymbol{z}_i^1$ and $\boldsymbol{z}_i^2$ represent the latent embedding of $i$-th frame in $\boldsymbol{V}^1$ and $\boldsymbol{V}^2$ respectively. Let $S^1=\{s_i^1~|~1\leq i \leq T\}$ denote timestamp vector of $\boldsymbol{V}^1$, where $s_i^1$ is the corresponding raw video timestamp of the $i$-th frame in $\boldsymbol{V}^1$ (see Figure~\ref{fig4}). In the same way, we can define $S^2=\{s_i^2~|~1\leq i \leq T\}$.

Given the $i$-th reference frame in $\boldsymbol{V}^1$ and its corresponding latent embedding $\boldsymbol{z}_i^1$, due to the fact that temporally adjacent frames are more highly correlated than those far-away ones, we assume the embedding similarity between $\boldsymbol{z}_i^1$ and $\boldsymbol{Z}^2=\{\boldsymbol{z}_i^2~|~1\leq i \leq T\}$ should follow a prior Gaussian distribution of timestamp distance between $s_i^1$ and $S^2=\{s_i^2~|~1\leq i \leq T\}$. This assumption motivates us to use KL-divergence to optimize the embedding space. Specifically, let $\operatorname{sim}(\boldsymbol{u}, \boldsymbol{v})=\boldsymbol{u}^{\top} \boldsymbol{v} /\|\boldsymbol{u}\|\|\boldsymbol{v}\|$ denote cosine similarity, and $\textit{G}(x)=\frac{1}{\sigma \sqrt{2 \pi}} \exp(-\frac{x^{2}}{2 \sigma^{2}})$ denote the Gaussian function, where $\sigma^2$ is the variance. We  formulate the loss of $i$-th reference frame in $\boldsymbol{V}^1$ as follows:
\begin{align}
    \mathcal{L}^1_i=-\sum_{j=1}^T & w_{ij} \log\frac{\exp(\operatorname{sim}(\boldsymbol{z}_i^1, \boldsymbol{z}_j^2)/\tau)}{\sum_{k=1}^T \exp(\operatorname{sim}(\boldsymbol{z}_i^1, \boldsymbol{z}_k^2)/\tau)},\\
    w_{ij}&=
    \frac{\textit{G}(s_i^1-s_j^2)}{\sum_{k=1}^T \textit{G}(s_i^1-s_k^2)}, \label{eq:contrastive_loss}
\end{align}
where $w_{ij}$ is the normalized Gaussian weight and $\tau$ is the temperature parameter. Then the overall loss for $\boldsymbol{V}^1$ can be computed across all frames:
\begin{align}
    \mathcal{L}^1&=\frac{1}{T}\sum_{i=1}^{T}\mathcal{L}^1_{i}. 
\end{align}
Similarly, we can calculate the loss $\mathcal{L}^2$ for $\boldsymbol{V}^2$. Our sequence contrastive loss is defined as $\mathcal{L}_{\text{SCL}} = \mathcal{L}^1 + \mathcal{L}^2$. Noticeably, our loss does not rely on frame-to-frame correspondence between $\boldsymbol{V}^1$ and $\boldsymbol{V}^2$, which supports the diversity of spatial-temporal data augmentation.

\section{Experiments}

\subsection{Datasets and Metrics}
We use three video datasets, namely \textit{PennAction}~\cite{PennAction}, \textit{FineGym}~\cite{FineGym} and \textit{Pouring}~\cite{TCN} to evaluate the performance of our method. We compare our method with sate-of-the-arts on all three datasets. Unless otherwise specified, all ablation studies on conducted on PennAction dataset. 

\noindent\textbf{PennAction Dataset.} Videos in this dataset show humans doing different kinds of sports or exercise. Following TCC~\cite{TCC}, we use 13
actions of PennAction dataset. In total, there are 1140 videos for training and 966 videos for testing. Each action set has 40-134 videos for training and 42-116 videos for testing. We obtain per-frame labels from LAV~\cite{LAV}. The video frames are from 18 to 663.


\noindent\textbf{FineGym Dataset.} FineGym is a recent large-scale fine-grained action recognition dataset that requires representation learning methods to distinguish different sub-actions within the same video. We chunk the original YouTube videos according to the action boundaries so that each trimmed video data only describes a single action type (Floor Exercise, Balance Beam, Uneven Bars, or Vault-Women). Finally, we obtained 3182 videos for training and 1442 videos for testing. The video frames vary from 140 to 5153. FineGym provides two data splits according to the category number, namely FineGym99 with 99 sub-action classes and FineGym288 with 288 sub-action classes.

\noindent\textbf{Pouring Dataset.} In this dataset, videos record the process of hand pouring water from one object to another. The phase labels (5 phase classes) are obtained from TCC~\cite{TCC}. Following TCC~\cite{TCC}, we use 70 videos for training and 14 videos for testing. The video frames are from 186 to 797. 
\vspace{+0.3cm}
\begin{table*}[t]
	\centering
	\begin{tabular}{l|c|c|c|c|c}
		\toprule
		Method & Training Strategy & Annotation & Classification & Progress & $\tau$ \\
		\midrule
		TCC~\cite{TCC} & \multirow{2}{*}{Per-action} & \multirow{2}{*}{Weakly} & 81.35 & 0.664 & 0.701 \\
		LAV~\cite{LAV} & & & 84.25 & 0.661 & 0.805 \\
		\midrule
		TCC~\cite{TCC} & \multirow{3}{*}{Joint} & \multirow{3}{*}{Weakly} & 74.39 & 0.591 & 0.641 \\
		LAV~\cite{LAV} & & & 78.68 & 0.625 & 0.684\\
		GTA~\cite{GTA} & & & - & 0.789 & 0.748 \\
		\midrule
		SaL~\cite{SAL} & \multirow{3}{*}{Joint} & \multirow{3}{*}{None} & 68.15 & 0.390 & 0.474 \\
		TCN~\cite{TCN} & & & 68.09 & 0.383 & 0.542 \\
		Ours & & & \textbf{93.07}  & \textbf{0.918}  & \textbf{0.985}  \\
		\bottomrule
	\end{tabular}
	\vspace{-2mm}
	\caption{Comparison with state-of-the-art methods on PennAction, using various evaluation metrics: \textit{Phase Classification} (Classification), \textit{Phase Progression} (Progress) and \textit{Kendall’s Tau} ($\tau$). The top row results are from per-action models, i.e., separate models are trained for different actions. The results in middle and bottom row are obtained from training a single model for all actions.}
	\vspace{-3mm}
	\label{tab2}
\end{table*}

\noindent\textbf{Evaluation Metrics.} For each dataset, We first optimize our network on the training set, without using any labels, and then use the following four metrics to evaluate the frame-wise representations: \begin{itemize}
\item \textit{Phase Classification} (or \textit{Fine-grained Action Classification})~\cite{TCC} is the averaged per-frame classification accuracy on testing set. Before testing, we fix the network and train a linear classifier by using per-frame labels (phase class or sub-action category) of the training set.
\item \textit{Phase Progression}~\cite{TCC} measures the representation ability to predict the phase progress. We fix the network and train a linear regressor to predict the phase progression values (timestamp distance between a query frame and phase boundaries) for all frames. Then it is computed as the average R-squared measure.
\item \textit{Kendall's Tau}~\cite{TCC} is calculated over every pair of testing videos by sampling two frames in the first video and retrieving the corresponding nearest frames in the second video, and checking whether their orders are shuffled. It measures how well-aligned two sequences are in time. No more training or finetuning is needed.
\item \textit{Average Precision@K}~\cite{LAV} is computed as how many frames in the retrieved K frames have the same phase labels as the query frame. It measures the fine-grained frame retrieval accuracy. $K=5, 10, 15$ are evaluated. No more training or finetuning is needed.
\end{itemize}

Following~\cite{TCC,TCN,LAV}, \textit{Phase Classification}, \textit{Phase Progression} and \textit{Kendall’s Tau} are evaluated on Pouring dataset. For PennAction, all four metrics are evaluated within each action category, and the final results are averaged across the 13 action categories. Following~\cite{GTA}, we use \textit{Fine-grained Action Classification} to evaluate our method on FineGym dataset.

\subsection{Implementation Details}
In our network, we adopt ResNet-50~\cite{ResNet} pre-trained by BYOL~\cite{BYOL} as frame-wise spatial encoder. Unless otherwise specified, we use a 3-layer Transformer encoder~\cite{Transformer} with 256 hidden size and 8 heads to model temporal context. We train the model using Adam optimizer with learning rate $10^{-4}$ and weight decay $10^{-5}$. We decay the learning rate with cosine decay schedule without restarts~\cite{cosine}. In our loss, we set $\sigma^2=10$ and $\tau=0.1$ as default. Following SimCLR~\cite{SimCLR}, random image cropping, horizontal flipping, random color distortions, and random Gaussian blur are employed as the spatial augmentations. For our temporal data augmentations described in Section~\ref{sec:view}, we set hyper-parameters $\alpha=1.5$ and $\beta=20\%$. The video batch size is set as 4 (8 views), and our model is trained on 4 Nvidia V100 GPUs for 300 epochs. During training, we sample $T=240$ frames for Pouring and FineGym, $T=80$ frames for PennAction. During testing, we feed the whole video into the model at once, without any temporal down-sampling. We L2-normalize the frame-wise representations for evaluation.

\subsection{Main Results}
\paragraph{Results on PennAction Dataset.} In Table~\ref{tab2}, our method is compared with state-of-the-art methods on PennAction. TCC~\cite{TCC} and LAV~\cite{LAV} train a separate model for each action (`Per-action' in the table), which results in 13 expert models for 13 action classes correspondingly. In contrast, we train only one model for all 13 action classes (`Joint' in the table). Noticeably, our approach not only outperforms the methods using joint training, but also outperforms the methods adopting per-action training strategy by a large margin under different evaluation metrics. In Table~\ref{tab2-2}, we report the results under the \textit{Average Precision@K} metric, which measures the performance of fine-grained frame retrieval. Surprisingly, although our model is not trained on paired data, it can successfully find frames with similar semantics from other videos. For all AP@K, our method is at least +11\% better than previous methods. 
\vspace{+0.3cm}

\begin{table}
	\centering
	\begin{tabular}{l|c|c|c}
		\toprule
		Method & AP@5 & AP@10 & AP@15 \\
		\midrule
		TCN~\cite{TCN} & 77.84 & 77.51 & 77.28 \\
		TCC~\cite{TCC} & 76.74 & 76.27 & 75.88 \\
		LAV~\cite{LAV} & 79.13 & 78.98 & 78.90 \\
		\midrule
		Ours & \textbf{92.28}  & \textbf{92.10}  & \textbf{91.82}  \\
		\bottomrule
	\end{tabular}
	\vspace{-2mm}
	\caption{Fine-grained frame retrieval results on PennAction.}
	\vspace{-3mm}
	\label{tab2-2}
\end{table}

\begin{table}
	\centering
	\begin{tabular}{l|c|c}
		\toprule
		Method & FineGym99 & FineGym288 \\
		\midrule
		$\mathrm{D}^{3} \mathrm{TW}$~\cite{D3TW} & 15.28 & 14.07 \\
		SpeedNet~\cite{SpeedNet} & 16.86 & 15.57 \\
		TCN~\cite{TCN} & 20.02 & 17.11 \\
		SaL~\cite{SAL} & 21.45 & 19.58 \\
		TCC~\cite{TCC} & 25.18 & 20.82 \\
		GTA~\cite{GTA} & 27.81 & 24.16 \\
		\midrule
		Ours & \textbf{41.75}  &  \textbf{35.23} \\
		\bottomrule
	\end{tabular}
	\vspace{-2mm}
	\caption{Comparison with state-of-the-art methods on FineGym, under the evaluation of \textit{Fine-grained Action Classification}.}
	\vspace{-6mm}
	\label{tab1}
\end{table}

\noindent\textbf{Results on FineGym Dataset.} Table~\ref{tab1} summarizes the experimental results of \textit{Fine-grained Action Classification} on FineGym99 and FineGym288. Our method outperforms the other self-supervised~\cite{SpeedNet, TCN, SAL} and weakly supervised~\cite{TCC, D3TW, GTA} methods. The performance of our method surpasses the previous state-of-the-art method GTA~\cite{GTA} by +13.94\% on FineGym99 and +11.07\% on FineGym288. The weakly supervised methods, i.e., TCC~\cite{TCC}, $\mathrm{D}^{3} \mathrm{TW}$~\cite{D3TW} and GTA~\cite{GTA}, assume there exists an optimal alignment between two videos from the training set. However, for FineGym dataset, even in two videos describing the same action, the set and order of sub-actions may differ. Therefore, the alignment found by these methods can be incorrect, which impedes learning. The great improvement verifies the effectiveness of our framework.
\vspace{+0.3cm}

\noindent\textbf{Results on Pouring Dataset.} As shown in Table~\ref{tab3}, our method also achieves the best performance on a relatively small dataset, Pouring. These results further demonstrate the great generalization ability of our approach.
\vspace{+0.3cm}

\noindent\textbf{Visualization Results.} We present the visualization of fine-grained frame retrieval and video alignment in Section A.

\begin{table}
  \centering
  \begin{tabular}{l|c|c|c}
    \toprule
    Method & Classification & Progress & $\tau$ \\
    \midrule
    TCN~\cite{TCN} & 89.53 & 0.804 & 0.852 \\
    TCC~\cite{TCC} & 91.53 & 0.837 & 0.864 \\
    LAV~\cite{LAV} & 92.84 & 0.805 & 0.856 \\
    \midrule
    Ours & \textbf{93.73}  & \textbf{0.935}  & \textbf{0.992}  \\
    \bottomrule
  \end{tabular}
  \vspace{-2mm}
  \caption{Comparison with state-of-the-art methods on Pouring.}
  \label{tab3}
\end{table}

\begin{table}
	\centering
	\begin{tabular}{l|c|c|c}
		\toprule
		Architecture & Classification & Progress & $\tau$ \\
		\midrule
		ResNet-50 only & 68.63 & 0.296 & 0.440 \\
		ResNet-50+C3D & 83.96 & 0.705 & 0.778 \\
		\midrule
		\textbf{ResNet-50+} & \multirow{2}{*}{\textbf{93.07}}  & \multirow{2}{*}{\textbf{0.918}}  & \multirow{2}{*}{\textbf{0.985}} \\
		\textbf{Transformer} &  &  & \\
		\bottomrule
	\end{tabular}
	\vspace{-2mm}
	\caption{Ablation study on different architectures.}
	\vspace{-4mm}
	\label{tab4}
\end{table}

\subsection{Ablation Study}
In this section, we perform multiple experiments to analyze the different components of our framework. Unless otherwise specified, experiments are conducted on the \textit{PennAction} dataset.

\noindent\textbf{Network Architecture.} In Table~\ref{tab4}, we investigate the network architecture. `ResNet-50+Transformer' denotes our default frame-level video encoder introduced in Section~\ref{sec:encoder}. `ResNet-50 only' means we remove the Transformer encoder in our network, and only use 2D ResNet-50 and linear transformation layers to extract representations per frame. `ResNet-50+C3D' represents that two 3D convolutional layers~\cite{C3D} are added on top of the ResNet-50 before the spatial pooling, which is the same as the model adopted in TCC~\cite{TCC} and LAV~\cite{LAV}. These models are all trained with the proposed sequence contrastive loss. Our default network outperforms the other two networks, which attributes to the long-range dependency modeling ability of Transformers.

\noindent\textbf{Layer Number of Transformer Encoder.} Table~\ref{tab5} shows studies using different numbers of layers in Transformers. We find that \textit{Phase Classification} increases with more layers. However, \textit{Phase Progression} slightly drops when there are too many layers. We use 3 layers by default.

\begin{table}
	\centering
	\begin{tabular}{c|c|c|c}
		\toprule
		\#Layers & Classification & Progress & $\tau$ \\
		\midrule
		1 & 92.15 & 0.909 & 0.985 \\
		2 & 92.61 & 0.913 & \textbf{0.990} \\
		\textbf{3} & \textbf{93.07}  & \textbf{0.918}  & 0.985 \\
		4 & 92.81 & 0.910 & \textbf{0.990} \\
		\bottomrule
	\end{tabular}
	\vspace{-2mm}
	\caption{Study on the effects of using different number of layers in Transformer encoder.}
	\label{tab5}
\end{table}

\begin{table}
	\centering
	\begin{tabular}{l|c|c|c}
		\toprule
		Learnable Blocks & Classification & Progress & $\tau$ \\
		\midrule
		None & 90.63 & 0.907 & \textbf{0.994} \\
		\textbf{Block5} & \textbf{93.07}  & 0.918  & 0.985 \\
		Block4+Block5 & 92.98 & \textbf{0.919} & 0.989 \\
		\bottomrule
	\end{tabular}
	\vspace{-2mm}
	\caption{Ablation study on learnable blocks of ResNet-50.}
	\label{tab6}
\end{table}

\begin{table}
	\centering
	\begin{tabular}{l|c|c|c}
		\toprule
		Method & Classification & Progress & $\tau$ \\
		\midrule
		TCN$^\dagger$ & 86.31 & 0.898 & 0.832 \\
		TCC$^\dagger$ & 86.35 & 0.899 & 0.980 \\
		\midrule
		Ours & \textbf{93.07} & \textbf{0.918} & \textbf{0.985} \\
		\bottomrule
	\end{tabular}
	\vspace{-2mm}
	\caption{Applying our network to TCN and TCC. $^\dagger$ denotes we re-implement the method and replace the network with ours. ``Contrastive baseline'' uses the corresponding frame at the other view as the positive sample. }
	\vspace{-4mm}
	\label{tab7}
\end{table}

\noindent\textbf{Training Different Blocks of ResNet.}
In our implementation, ResNet-50 is pre-trained on ImageNet. In Table~\ref{tab6}, we study the effects of finetuning different blocks of ResNet-50. The standard ResNet contains 5 blocks, namely \textit{Block1-Block5}. `None' denotes that all layers of ResNet are frozen. `Block5' denotes we freeze the first four residual blocks of ResNet and only make the last residual block learnable, which is our default setting. Similarly, `Block4+Block5' means we freeze the first three blocks and only train the last two blocks. Table~\ref{tab6} shows that encoding dataset-related spatial information is important (`None' vs. `Block5'), and training more blocks does not lead to improvement (`Block5' vs. `Block4+Block5').

\noindent\textbf{Applying Our Network to Other Methods.}
We study whether our frame-level video encoder (FVE) introduced in Section~\ref{sec:encoder} can boost the performances of TCC~\cite{TCC} and TCN~\cite{TCN}. We replace the C3D-based network with ours. Table~\ref{tab7} shows the results. We find that the proposed network can dramatically improve the performance of their methods (compared with the results in Table~\ref{tab1}). In addition, our method still keeps a large performance gain, which attributes to the proposed sequence contrastive loss.

\begin{table}
	\centering
	\begin{tabular}{c|c|c|c}
		\toprule
		 Hyper-parameters & Classification & Progress & $\tau$ \\
		\midrule
		$\tau$=0.1, $\sigma^2$=1 & 92.95 & 0.903 & 0.963 \\
		$\tau$=0.1, $\sigma^2$=25 & 92.03 & \textbf{0.922} & \textbf{0.993} \\
		\midrule
		$\tau$=1.0, $\sigma^2$=10 & 91.57 & 0.889 & 0.993 \\
		$\tau$=0.3, $\sigma^2$=10 & 92.13 & 0.903 & 0.992 \\
		$\tau$=0.1, $\sigma^2$=10 & \textbf{93.07} & 0.918 & 0.985 \\
		\bottomrule
	\end{tabular}
	\vspace{-2mm}
	\caption{Ablation study on Gaussian variance $\sigma^2$ and the temperature $\tau$ in sequence contrastive loss.}
	\vspace{-2mm}
	\label{tab8}
\end{table}

\noindent\textbf{Hyper-parameters of Sequence Contrastive Loss.}
We study the hyper-parameters, i.e., temperature parameter $\tau$ and Gaussian variance $\sigma^2$ in our sequence contrastive loss (see Eq.~\ref{eq:contrastive_loss}). The variance $\sigma^2$ of the prior Gaussian distribution controls how the adjacent frames are semantically similar to the reference frame, on the assumption. As Table~\ref{tab8} shows, too small variance ($\sigma^2=1$) or too large variance ($\sigma^2=25$) degrades the performance. We use $\sigma^2=10$ by default. In addition, we observe an appropriate temperature ($\tau=0.1$) facilitates the learning from hard negatives, which is consistent with the conclusion in SimCLR~\cite{SimCLR}.

\noindent\textbf{Study on Different Temporal Data Augmentations.}
We study the different temporal data augmentations described in Section~\ref{sec:view}, including maximum crop size $\alpha$, overlap ratio $\beta$ between views, and different sampling strategies, namely random sampling and even sampling. Table~\ref{tab9} shows the results. From the table, we can see that the performance drops dramatically when we crop the video with a fixed length ($\alpha=1$). The performance also decreases when we perform even sampling on the cropped clips. As described in Section~\ref{sec:loss}, our sequence contrastive loss does not rely on frame-to-frame correspondence between two augmented views. Experimentally, constructing two views with $\beta=100\%$ percent of overlapped frames degrades the performance, since the variety of augmented data decreases. In addition, we also observe the performance drops when two views are constructed independently ($\beta=0\%$ ). The reason is that in this setting, the training may bring the representations of temporally distant frames closer, which hinders the optimization.

\begin{table}
	\centering
	\begin{tabular}{c|c|c|c}
		\toprule
		$\alpha$ & Sampling & $\beta$ (\%) & FineGym99 \\
		\midrule
		0 & \multirow{3}{*}{Random} & \multirow{3}{*}{20} & 36.72 \\
		\textbf{1.5} & & &\textbf{41.75} \\
		1 & & &39.03 \\
		\midrule
		1.5 & Even & 20 & 38.44 \\
		\midrule
		\multirow{5}{*}{1.5} & \multirow{5}{*}{Random} & 0 & 38.15 \\
		 & & \textbf{20} & \textbf{41.75} \\
		 & & 50 & 39.14 \\
		 & & 80 & 37.94 \\
		 & & 100 & 35.53 \\
		\bottomrule
	\end{tabular}
	\vspace{-2mm}
	\caption{Ablation study on hyper-parameters of temporal data augmentations. Effects of maximum crop size $\alpha$, overlap ratio $\beta$ and random sampling strategy are studied. The experiments are conducted on FineGym99 dataset.}
	\label{tab9}
\end{table}

\begin{table}
	\centering
	\begin{tabular}{c|c c c}
	    \toprule
        \% of Labeled Data $\to$ & 10 & 50 & 100 \\
		\midrule
		\multicolumn{4}{l}{\textit{Number of training frames:}} \\
		80 & 27.10  & 32.78  & 34.02 \\
		160 & 30.28 & 36.46 & 38.06 \\
		\textbf{240} & \textbf{33.53} & 39.89 & 41.75 \\
		480 & 31.46 & 37.92 & 39.45 \\
		\midrule
        Supervised & 24.51  & \textbf{48.75}  & \textbf{60.37} \\
		\bottomrule
	\end{tabular}
	\vspace{-2mm}
	\caption{Ablation studies on number of training frames under different data protocols. Study is conducted on FineGym99 \textit{Fine-grained Action Classification} task. `Supervised' means all layers are trained with supervised learning.}
	\vspace{-4mm}
	\label{tab10}
\end{table}

\noindent\textbf{Number of Training Frames and Linear Evaluation Under Different Data Protocols.}
As described in Section~\ref{sec:view}, our network takes augmented views with $T$ frames as input. We study the effects of different frame numbers $T$ on FineGym99. Table~\ref{tab10} shows the results. We observe that taking long sequences as input is essential for frame-wise representation learning. However, a too large frame number degrades the performance. We thus set $T=240$ by default. We also conduct linear evaluation under different data protocols. Concretely, we use $10\%$, $50\%$ and $100\%$ labeled data to train the linear classifier. Compared with the supervised model (all layers are learnable), our method achieves better performance when the labeled data is limited ($10\%$ data protocol).

\section{Conclusion}
In this paper, we present a novel framework named contrastive action representation learning (CARL) to learn frame-wise action representations, especially for long videos, in a self-supervised manner. To model long videos with hundreds of frames, we introduce a simple yet efficient network named frame-level video encoder (FVE), which considers spatio-temporal context during training. In addition, we propose a novel sequence contrastive loss (SCL) for frame-wise representation learning. SCL optimizes the embedding space by minimizing the KL-divergence between the  sequence similarity of two augmented views and a prior Gaussian distribution. Experiments on various datasets and tasks show effectiveness and generalization of our method.

\subsection*{Acknowledgments}

This work was supported in part by The National Key Research and Development Program of China (Grant Nos: 2018AAA0101400), in part by The National Nature Science Foundation of China (Grant Nos: 62036009, 61936006), in part by Innovation Capability Support Program of Shaanxi (Program No. 2021TD-05).

\appendix

\begin{figure}[t]
    \centering
        \centering
        \includegraphics[width=0.98\linewidth]{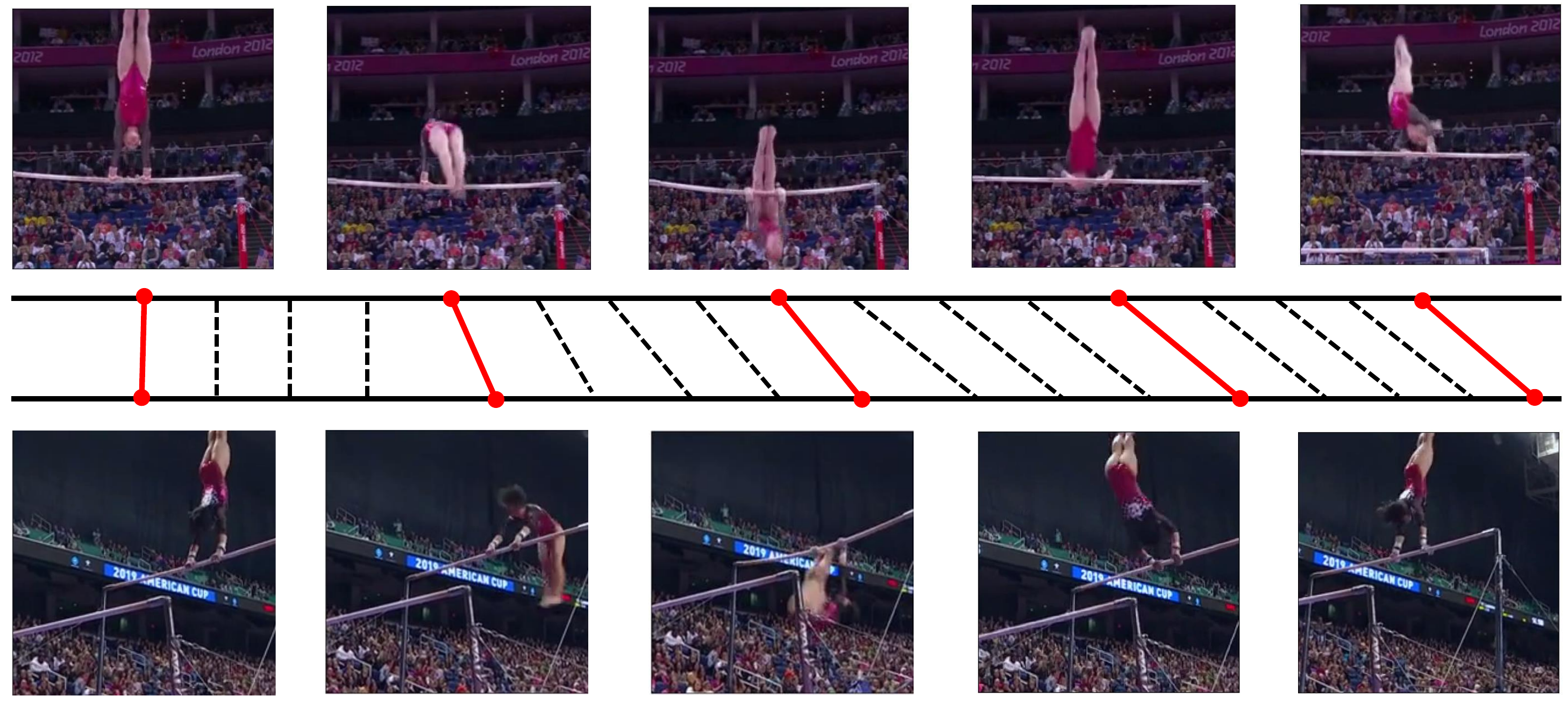}
    \caption{Visualization of video alignment on FineGym dataset. Please refer to video demos in our supplementary materials for more visualization results.}
    \vspace{-0.2cm}
    \label{supp_fig1}
\end{figure}

\section{More Results}
In this section, we show visualization results of video alignment and fine-grained frame retrieval.

\begin{figure}[t]
  \centering
  \begin{subfigure}[]{0.49\linewidth}
    \centering
    \includegraphics[width=\linewidth]{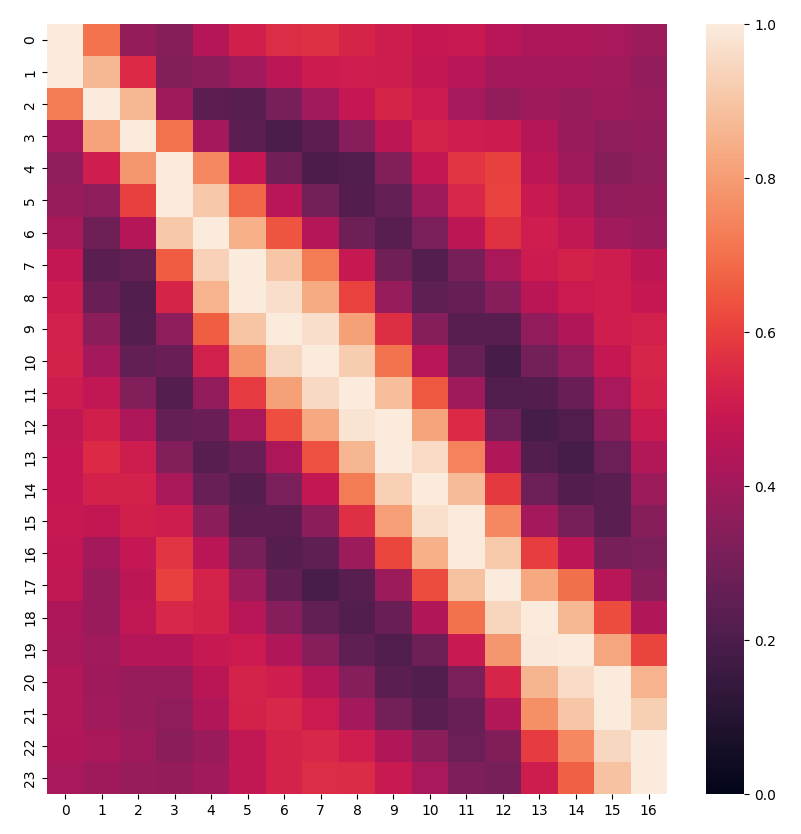}
    \caption{Pouring dataset.}
  \end{subfigure}
  \begin{subfigure}[]{0.49\linewidth}
    \centering
    \includegraphics[width=\linewidth]{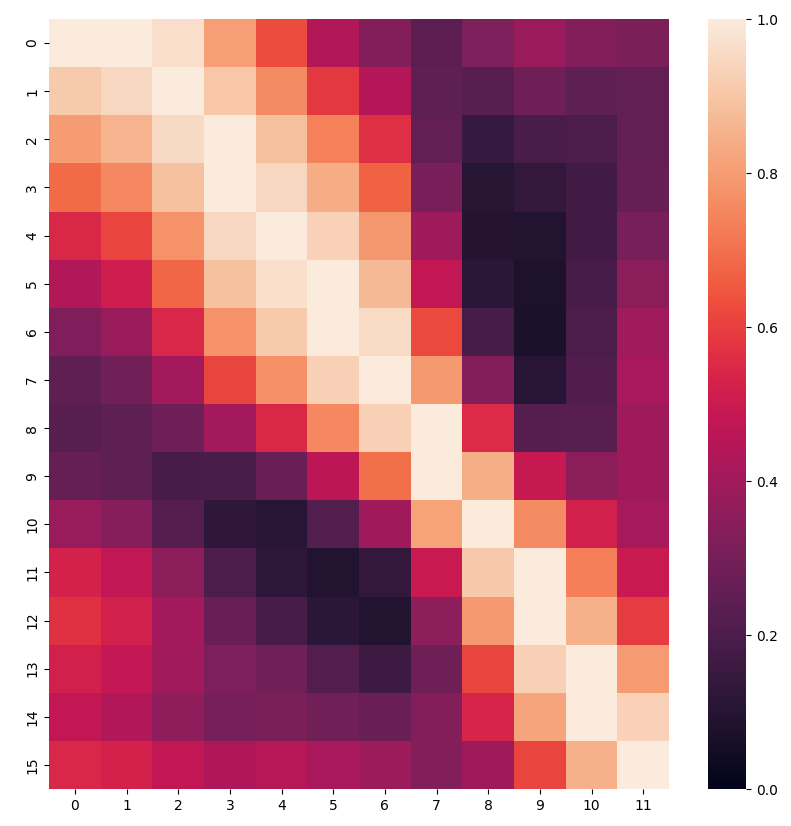}
    \caption{PennAction dataset.}
  \end{subfigure}
  \vspace{-0.2cm}
  \caption{We randomly select two videos recording the same process (or action) from Pouring (or PennAction) dataset and compute the similarity matrix for frame-wise representations extracted by our method. The similarities are normalized for better visualization.}
   \vspace{-0.2cm}
   \label{supp_fig2}
\end{figure}

\begin{figure}[t]
    \centering
    \includegraphics[width=0.98\linewidth]{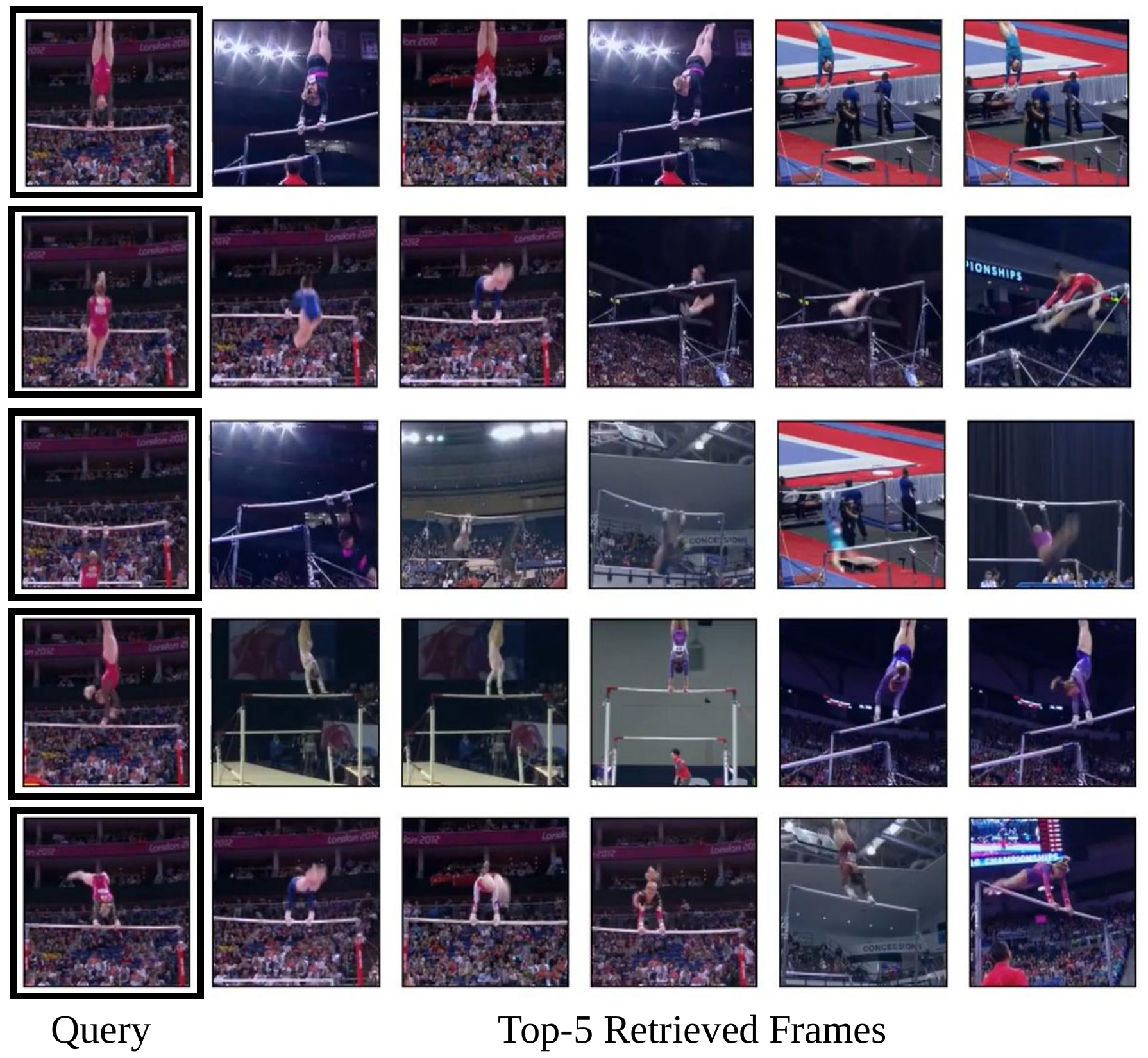}
    \caption{Visualization of fine-grained frame retrieval on FineGym datast by using our method. 
    }
    \vspace{-0.2cm}
    \label{supp_fig3}
\end{figure}

\subsection{Video Alignment}

Given two videos recording the similar action or process, the goal of video alignment is to find the temporal correspondence between them. Firstly, we use our framework to extract the frame-wise representations for two randomly selected videos. Then we compute the cosine similarities between the frame-wise representations of two videos and utilize the famous dynamic time warping (DTW) algorithm on the similarity matrix to find the best temporal alignment. Figure~\ref{supp_fig1} shows an example from FineGym test set. Please refer to video demos in our supplementary materials for more visualization results.

We also randomly select two videos recording the same process (or action) from Pouring (or PennAction) dataset, and similarly, we can compute the similarity matrix which is rendered as a heatmap in Figure~\ref{supp_fig2}. We observe that the diagonal is highlighted, which means our approach find the favorable alignment between two correlated videos. We also give video demos in our supplementary materials.

\subsection{Fine-grained Frame Retrieval}

In Figure~\ref{supp_fig3}, we present the visualization results of fine-grained frame retrieval on FineGym dataset. To be specific, we feed the video containing the query frames into our CARL framework to generate query features, and similarly, we can extract frame-wise features for the rest videos in the test set. We simply compute the cosine similarity between query features and frame-wise features from candidate videos to obtain top-5 retrieved frames as shown in Figure~\ref{supp_fig3}. The retrieved frames have similar semantics with the query frame, though the appearances, the camera views, and the backgrounds are different, which suggests our method is robust to these factors.

\subsection{Action Localization}
To show the potential of our method on large datasets and more downstream tasks, we optimize the frame-wise features via our self-supervised method on ActivityNet~\cite{ActivityNet}. Then we use G-TAD~\cite{gtad} on the top of the features (without fine-tuning) to perform temporal action localization. As shown in Table~\ref{supp_tab11}, we use mAP(\%) at \{0.5, 0.75, 0.95\} tIoU thresholds and the average mAP across 10 tIoU levels for evaluation. In contrast to the supervised two-stream model~\cite{C3D}, our method does not need any video labels while achieving better performance.

\begin{table}
  \centering
  \begin{tabular}{l|c|c|c|c}
    \toprule
    Method & 0.5 & 0.75 & 0.95 & Average \\
    \midrule
    G-TAD w. 2stream & 50.36 & 34.60 & 9.02 & 34.09 \\
    G-TAD w. ours & 51.22 & 35.19 & 8.54 & 34.46 \\
    \bottomrule
  \end{tabular}
  \caption{Temporal action localization on ActivityNet v1.3.} 
  \label{supp_tab11}
\end{table}

\subsection{Compare with Contrastive Baseline}
We compare our SCL with the contrastive baseline which only uses the corresponding frame in the other view as the positive sample and ignores temporal adjacent frames. As Table~\ref{tab_diff_loss} shows, our SCL can more efficiently employ the sequential information and thus achieves better performance.

\begin{table}
	\centering
	\begin{tabular}{l|c|c|c}
		\toprule
		Method & Classification & Progress & $\tau$ \\
		\midrule
		Contrastive baseline & 88.05 & 0.898 & 0.891 \\
		SCL (ours) & \textbf{93.07} & \textbf{0.918} & \textbf{0.985} \\
		\bottomrule
	\end{tabular}
	\caption{Compare our SCL with contrastive baseline, which uses the corresponding frame in the other view as the positive sample.}
	\label{tab_diff_loss}
\end{table}

\begin{table}
	\centering
	\begin{tabular}{l|c|c|c}
		\toprule
		Training Dataset & Classification & Progress & $\tau$ \\
		\midrule
		K400  & 91.9 & 0.903 & 0.949 \\
		K400 $\rightarrow$ PennAction  & \textbf{93.9} & \textbf{0.908} & \textbf{0.977} \\
		\bottomrule
	\end{tabular}
	\caption{Our CARL pre-trained on Kinetics-400 shows outstanding transfer ability on PennAction. Fine-tuning the pre-trained model on PennAction further boosts the performance.}
	\label{tab_k400}
\end{table}

\subsection{Kinetics-400 Pre-training}
To show our method can benefit from large-scale datasets without any labels, we train our CARL on Kinetics-400~\cite{Kinetics}. As Table~\ref{tab_k400} shows, the frame-wise representations trained on Kinetics-400 shows outstanding generalization on PennAction dataset. Moreover, fine-tuning the pre-trained model on PennAction by using our CARL further boosts the performance, e.g., + 2\% classification improvement.

{\small
\bibliographystyle{ieee_fullname}
\bibliography{egbib}

\begin{thebibliography}{10}\itemsep=-1pt

\bibitem{ViViT}
Anurag Arnab, Mostafa Dehghani, Georg Heigold, Chen Sun, Mario Lucic, and
  Cordelia Schmid.
\newblock Vivit: A video vision transformer.
\newblock {\em ArXiv}, 2021.

\bibitem{SpeedNet}
Sagie Benaim, Ariel Ephrat, Oran Lang, Inbar Mosseri, William~T. Freeman,
  Michael Rubinstein, Michal Irani, and Tali Dekel.
\newblock Speednet: Learning the speediness in videos.
\newblock In {\em CVPR}, 2020.

\bibitem{TimeSformer}
Gedas Bertasius, Heng Wang, and Lorenzo Torresani.
\newblock Is space-time attention all you need for video understanding?
\newblock {\em ArXiv}, 2021.

\bibitem{camgoz2018neural}
Necati~Cihan Camgoz, Simon Hadfield, Oscar Koller, Hermann Ney, and Richard
  Bowden.
\newblock Neural sign language translation.
\newblock In {\em CVPR}, 2018.

\bibitem{camgoz2020sign}
Necati~Cihan Camgoz, Oscar Koller, Simon Hadfield, and Richard Bowden.
\newblock Sign language transformers: Joint end-to-end sign language
  recognition and translation.
\newblock In {\em CVPR}, 2020.

\bibitem{TemporalAlignment}
Kaidi Cao, Jingwei Ji, Zhangjie Cao, C. Chang, and Juan~Carlos Niebles.
\newblock Few-shot video classification via temporal alignment.
\newblock In {\em CVPR}, 2020.

\bibitem{DETR}
Nicolas Carion, Francisco Massa, Gabriel Synnaeve, Nicolas Usunier, Alexander
  Kirillov, and Sergey Zagoruyko.
\newblock End-to-end object detection with transformers.
\newblock {\em ArXiv}, 2020.

\bibitem{SwAV}
Mathilde Caron, Ishan Misra, Julien Mairal, Priya Goyal, Piotr Bojanowski, and
  Armand Joulin.
\newblock Unsupervised learning of visual features by contrasting cluster
  assignments.
\newblock {\em ArXiv}, 2020.

\bibitem{Kinetics}
Jo{\~a}o Carreira and Andrew Zisserman.
\newblock Quo vadis, action recognition? a new model and the kinetics dataset.
\newblock In {\em CVPR}, 2017.

\bibitem{D3TW}
C. Chang, De-An Huang, Yanan Sui, Li Fei-Fei, and Juan~Carlos Niebles.
\newblock D3tw: Discriminative differentiable dynamic time warping for weakly
  supervised action alignment and segmentation.
\newblock In {\em CVPR}, 2019.

\bibitem{SimCLR}
Ting Chen, Simon Kornblith, Mohammad Norouzi, and Geoffrey~E. Hinton.
\newblock A simple framework for contrastive learning of visual
  representations.
\newblock In {\em ICML}, 2020.

\bibitem{MoCoV2}
Xinlei Chen, Haoqi Fan, Ross~B. Girshick, and Kaiming He.
\newblock Improved baselines with momentum contrastive learning.
\newblock {\em ArXiv}, 2020.

\bibitem{chen2022simple}
Yutong Chen, Fangyun Wei, Xiao Sun, Zhirong Wu, and Stephen Lin.
\newblock A simple multi-modality transfer learning baseline for sign language
  translation.
\newblock {\em arXiv preprint arXiv:2203.04287}, 2022.

\bibitem{ImageNet}
Jia Deng, Wei Dong, Richard Socher, Li{-}Jia Li, Kai Li, and Fei{-}Fei Li.
\newblock Imagenet: {A} large-scale hierarchical image database.
\newblock In {\em CVPR}, 2009.

\bibitem{ViT}
Alexey Dosovitskiy, Lucas Beyer, Alexander Kolesnikov, Dirk Weissenborn,
  Xiaohua Zhai, Thomas Unterthiner, Mostafa Dehghani, Matthias Minderer, Georg
  Heigold, Sylvain Gelly, Jakob Uszkoreit, and Neil Houlsby.
\newblock An image is worth 16x16 words: Transformers for image recognition at
  scale.
\newblock In {\em ICLR}, 2021.

\bibitem{TCC}
Debidatta Dwibedi, Yusuf Aytar, Jonathan Tompson, Pierre Sermanet, and Andrew
  Zisserman.
\newblock Temporal cycle-consistency learning.
\newblock In {\em CVPR}, 2019.

\bibitem{SlowFast}
Christoph Feichtenhofer, Haoqi Fan, Jitendra Malik, and Kaiming He.
\newblock Slowfast networks for video recognition.
\newblock In {\em ICCV}, 2019.

\bibitem{largescale}
Christoph Feichtenhofer, Haoqi Fan, Bo Xiong, Ross~B. Girshick, and Kaiming He.
\newblock A large-scale study on unsupervised spatiotemporal representation
  learning.
\newblock In {\em CVPR}, 2021.

\bibitem{Something-Something}
Raghav Goyal, Samira~Ebrahimi Kahou, Vincent Michalski, Joanna Materzynska,
  Susanne Westphal, Heuna Kim, Valentin Haenel, Ingo Fr{\"u}nd, Peter~N.
  Yianilos, Moritz Mueller-Freitag, Florian Hoppe, Christian Thurau, Ingo Bax,
  and Roland Memisevic.
\newblock The “something something” video database for learning and
  evaluating visual common sense.
\newblock In {\em ICCV}, 2017.

\bibitem{BYOL}
Jean-Bastien Grill, Florian Strub, Florent Altch'e, Corentin Tallec, Pierre~H.
  Richemond, Elena Buchatskaya, Carl Doersch, Bernardo~{\'A}vila Pires,
  Zhaohan~Daniel Guo, Mohammad~Gheshlaghi Azar, Bilal Piot, Koray Kavukcuoglu,
  R{\'e}mi Munos, and Michal Valko.
\newblock Bootstrap your own latent: A new approach to self-supervised
  learning.
\newblock {\em ArXiv}, 2020.

\bibitem{GTA}
Isma Hadji, Konstantinos~G. Derpanis, and Allan~D. Jepson.
\newblock Representation learning via global temporal alignment and
  cycle-consistency.
\newblock In {\em CVPR}, 2021.

\bibitem{DenseCode}
Tengda Han, Weidi Xie, and Andrew Zisserman.
\newblock Video representation learning by dense predictive coding.
\newblock In {\em ICCVW}, 2019.

\bibitem{LAV}
Sanjay Haresh, Sateesh Kumar, Huseyin Coskun, Shahram~Najam Syed, Andrey Konin,
  M.~Zeeshan Zia, and Quoc-Huy Tran.
\newblock Learning by aligning videos in time.
\newblock In {\em CVPR}, 2021.

\bibitem{ResNet}
Kaiming He, Xiangyu Zhang, Shaoqing Ren, and Jian Sun.
\newblock Deep residual learning for image recognition.
\newblock In {\em CVPR}, 2016.

\bibitem{ActivityNet}
Fabian~Caba Heilbron, Victor Escorcia, Bernard Ghanem, and Juan~Carlos Niebles.
\newblock Activitynet: A large-scale video benchmark for human activity
  understanding.
\newblock In {\em CVPR}, 2015.

\bibitem{VideoCL}
Haofei Kuang, Yi Zhu, Zhi Zhang, Xinyu Li, Joseph Tighe, S{\"o}ren
  Schwertfeger, C. Stachniss, and Mu Li.
\newblock Video contrastive learning with global context.
\newblock {\em ArXiv}, 2021.

\bibitem{Breakfast}
Hilde Kuehne, Ali~Bilgin Arslan, and Thomas Serre.
\newblock The language of actions: Recovering the syntax and semantics of
  goal-directed human activities.
\newblock In {\em CVPR}, 2014.

\bibitem{li2021towards}
Kenneth Li, Xiao Sun, Zhirong Wu, Fangyun Wei, and Stephen Lin.
\newblock Towards tokenized human dynamics representation.
\newblock {\em arXiv preprint arXiv:2111.11433}, 2021.

\bibitem{Liu2018ImitationFO}
Yuxuan Liu, Abhishek Gupta, P. Abbeel, and Sergey Levine.
\newblock Imitation from observation: Learning to imitate behaviors from raw
  video via context translation.
\newblock {\em 2018 IEEE International Conference on Robotics and Automation
  (ICRA)}, 2018.

\bibitem{cosine}
Ilya Loshchilov and Frank Hutter.
\newblock Sgdr: Stochastic gradient descent with warm restarts.
\newblock {\em arXiv: Learning}, 2017.

\bibitem{SAL}
Ishan Misra, C.~Lawrence Zitnick, and Martial Hebert.
\newblock Shuffle and learn: Unsupervised learning using temporal order
  verification.
\newblock In {\em ECCV}, 2016.

\bibitem{MomentsIT}
Mathew Monfort, Bolei Zhou, Sarah~Adel Bargal, Alex Andonian, Tom Yan, Kandan
  Ramakrishnan, Lisa~M. Brown, Quanfu Fan, Dan Gutfreund, Carl Vondrick, and
  Aude Oliva.
\newblock Moments in time dataset: One million videos for event understanding.
\newblock {\em PAMI}, 2020.

\bibitem{VTN}
Daniel Neimark, Omri Bar, Maya Zohar, and Dotan Asselmann.
\newblock Video transformer network.
\newblock {\em ArXiv}, 2021.

\bibitem{VideoConstrast}
Rui Qian, Tianjian Meng, Boqing Gong, Ming-Hsuan Yang, H. Wang, Serge~J.
  Belongie, and Yin Cui.
\newblock Spatiotemporal contrastive video representation learning.
\newblock In {\em CVPR}, 2021.

\bibitem{Cooking}
Marcus Rohrbach, Anna Rohrbach, Michaela Regneri, Sikandar Amin, Mykhaylo
  Andriluka, Manfred Pinkal, and Bernt Schiele.
\newblock Recognizing fine-grained and composite activities using hand-centric
  features and script data.
\newblock In {\em IJCV}, 2015.

\bibitem{TCN}
Pierre Sermanet, Corey Lynch, Yevgen Chebotar, Jasmine Hsu, Eric Jang, Stefan
  Schaal, and Sergey Levine.
\newblock Time-contrastive networks: Self-supervised learning from video.
\newblock {\em 2018 IEEE International Conference on Robotics and Automation
  (ICRA)}, 2018.

\bibitem{FineGym}
Dian Shao, Yue Zhao, Bo Dai, and Dahua Lin.
\newblock Finegym: A hierarchical video dataset for fine-grained action
  understanding.
\newblock In {\em CVPR}, 2020.

\bibitem{Charades}
Gunnar~A. Sigurdsson, G{\"u}l Varol, X. Wang, Ali Farhadi, Ivan Laptev, and
  Abhinav Gupta.
\newblock Hollywood in homes: Crowdsourcing data collection for activity
  understanding.
\newblock In {\em ECCV}, 2016.

\bibitem{TwoStream}
Karen Simonyan and Andrew Zisserman.
\newblock Two-stream convolutional networks for action recognition in videos.
\newblock In {\em NIPS}, 2014.

\bibitem{UCF101}
Khurram Soomro, Amir~Roshan Zamir, and Mubarak Shah.
\newblock Ucf101: A dataset of 101 human actions classes from videos in the
  wild.
\newblock {\em ArXiv}, 2012.

\bibitem{C3D}
Du Tran, Lubomir~D. Bourdev, Rob Fergus, Lorenzo Torresani, and Manohar Paluri.
\newblock Learning spatiotemporal features with 3d convolutional networks.
\newblock In {\em ICCV}, 2015.

\bibitem{Transformer}
Ashish Vaswani, Noam~M. Shazeer, Niki Parmar, Jakob Uszkoreit, Llion Jones,
  Aidan~N. Gomez, Lukasz Kaiser, and Illia Polosukhin.
\newblock Attention is all you need.
\newblock In {\em NIPS}, 2017.

\bibitem{TSN}
Limin Wang, Yuanjun Xiong, Zhe Wang, Yu Qiao, Dahua Lin, Xiaoou Tang, and
  Luc~Van Gool.
\newblock Temporal segment networks for action recognition in videos.
\newblock {\em PAMI}, 2019.

\bibitem{Nonlocal}
X. Wang, Ross~B. Girshick, Abhinav Gupta, and Kaiming He.
\newblock Non-local neural networks.
\newblock In {\em CVPR}, 2018.

\bibitem{wei2021aligning}
Fangyun Wei, Yue Gao, Zhirong Wu, Han Hu, and Stephen Lin.
\newblock Aligning pretraining for detection via object-level contrastive
  learning.
\newblock {\em Advances in Neural Information Processing Systems}, 34, 2021.

\bibitem{wu2018unsupervised}
Zhirong Wu, Yuanjun Xiong, Stella~X Yu, and Dahua Lin.
\newblock Unsupervised feature learning via non-parametric instance
  discrimination.
\newblock In {\em CVPR}, 2018.

\bibitem{gtad}
Mengmeng Xu, Chen Zhao, David~S. Rojas, Ali Thabet, and Bernard Ghanem.
\newblock G-tad: Sub-graph localization for temporal action detection.
\newblock In {\em CVPR}, 2020.

\bibitem{xu2021cross}
Yinghao Xu, Fangyun Wei, Xiao Sun, Ceyuan Yang, Yujun Shen, Bo Dai, Bolei Zhou,
  and Stephen Lin.
\newblock Cross-model pseudo-labeling for semi-supervised action recognition.
\newblock {\em arXiv preprint arXiv:2112.09690}, 2021.

\bibitem{SeCo}
Ting Yao, Yiheng Zhang, Zhaofan Qiu, Yingwei Pan, and Tao Mei.
\newblock Seco: Exploring sequence supervision for unsupervised representation
  learning.
\newblock In {\em AAAI}, 2021.

\bibitem{PennAction}
Weiyu Zhang, Menglong Zhu, and Konstantinos~G. Derpanis.
\newblock From actemes to action: A strongly-supervised representation for
  detailed action understanding.
\newblock In {\em ICCV}, 2013.

\end{thebibliography}
}

\end{document}